\documentclass[letterpaper, conference]{ieeeconf}
\IEEEoverridecommandlockouts
\usepackage{cite}

\usepackage{easyReview}
\usepackage{amsmath,amssymb,amsfonts}
\usepackage{algorithmic}
\usepackage{graphicx}
\usepackage{textcomp}
\usepackage{xcolor}

\begin{document}

\title{\LARGE \bf Soft Cap for Eversion Robots
}
\author{Cem Suulker$^{1}$,  Sophie Skach$^{1}$, Danyaal Kaleel$^{1}$, Taqi Abrar$^{2}$, Zain Murtaza$^{2}$, \\ Dilara Suulker$^{1}$, and Kaspar Althoefer$^{1}$, \it{Senior Member, IEEE}%
\thanks{This paper is accepted to and will be published at IEEE IROS 2023 conference.}
\thanks{For the purpose of open access, the author(s) has applied a Creative Commons Attribution (CC BY) license to any Accepted Manuscript version arising.}
\thanks{$^{1}$Authors are with the Centre for Advanced Robotics @ Queen Mary, School of Engineering and Materials Science, and $^{2}$School of Electronic Engineering and Computer Science, Queen Mary University of London, United Kingdom.
        {\tt\footnotesize c.suulker@qmul.ac.uk}}%
}
\maketitle

\begin{abstract}

Growing robots based on the eversion principle are known for their ability to extend rapidly, from within, along their longitudinal axis, and, in doing so, reach deep into hitherto inaccessible, remote spaces. Despite many advantages, eversion robots also present significant challenges, one of which is maintaining sensory payload at the tip without restricting the eversion process. A variety of tip mechanisms have been proposed by the robotics community, among them rounded caps of relatively complex construction that are not always compatible with functional hardware, such as sensors or navigation pouches, integrated with the main eversion structure. Moreover, many tip designs incorporate rigid materials, reducing the robot's flexibility and consequent ability to navigate through narrow openings. Here, we address these shortcomings and propose a design to overcome them: a soft, entirely fabric based, cylindrical cap that can easily be slipped onto the tip of eversion robots. Having created a series of caps of different sizes and materials, an experimental study was conducted to evaluate our new design in terms of four key aspects: eversion robot made from multiple layers of everting material, solid objects protruding from the eversion robot, squeezability, and navigability. In all scenarios, we can show that our soft, flexible cap is robust in its ability to maintain its position and is capable of transporting payloads such as a camera across long distances. We also demonstrate that the robot’s ability to move through restricted aperture openings and indeed its overall flexibility is virtually unhindered by the addition of our cap. The paper discusses the advantages of this design and gives further recommendations in relation to aspects of its engineering. 


\end{abstract}


\section{Introduction}

\begin{figure}[t]
  \centering
  \includegraphics[width=0.9\linewidth]{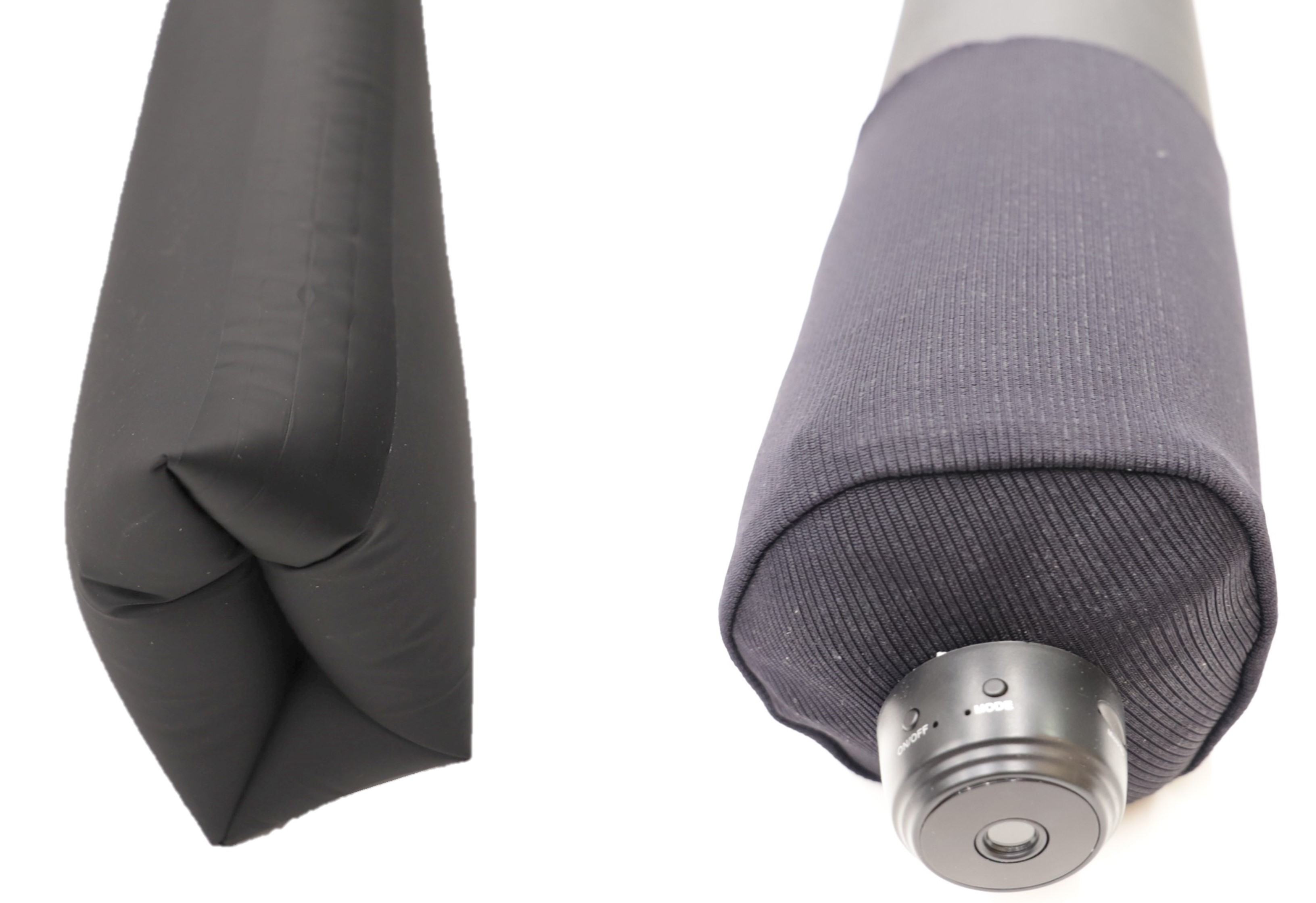}
  \caption{Left, an inflated eversion robot. Right, novel soft cap slipped over tip of eversion robot. A tool, here a camera, is attached to the cap.}
  \label{fig1}
\end{figure}
There is growing interest from a range of industries (prime examples being nuclear, construction, telecommunications, search and rescue, archaeology, and medicine) in robots that are capable of penetrating hard-to-access spaces, and conducting remote inspection, maintenance and repair tasks. The term `hard-to-access spaces' includes those that are physically constrained as well as those that present danger to humans, such as excessive nuclear radiation or the possibility of collapsing infrastructure. 
One of the key requirements for such tasks is the capability to travel some distance, along restricted channels, while carrying the tools with which to affect whatever tasks may be required.  
Small mobile robots have featured strongly in this context \cite{wheeledreview}, though many have proven ineffective as they can get lost, are difficult to recover if broken down, cannot easily overcome obstacles and,  when exposed to potentially hostile environments, can end up with damaged electronics and locomotion mechanisms \cite{mishima2006development}. 
An alternative route to overcoming the hard-to-access issue is provided by continuum robots. These snake-like robots, with a high length-to-diameter-ratio, can easily pass through small apertures and extend into the space behind \cite{webster2010design,sadati2017mechanics}. 
The development of the eversion robot - a new kind of continuum robot also known as a vine robot - is significant \cite{mishima2006development, hawkes2017soft}. In contrast to earlier continuum robots, eversion robots grow from the tip. Likened to the way plants grow, the cylindrical-shaped eversion robot made from an airtight fabric or polyethylene skin ejects its inner structure at the tip using pneumatic pressure - this is best imagined by considering a jacket's sleeve, detached from the jacket itself, continuously unfolding, the inside lining becoming the outer skin \cite{putzu2018plant}. The achieved motion is frictionless longitudinal growth (see Fig. 2) - a clear advantage in situations where the path is long or tortuous or the environment should be disturbed as little as possible \cite{vitanov2021suite}.  They can extend their length hundreds of times with regards to their folded state without putting pressure onto their surrounding environment and can bend passively in accordance with the environment, conforming to the shape of their surroundings \cite{godaba2019payload}.

\begin{figure*}[]
  \centering
  \includegraphics[width=1\textwidth]{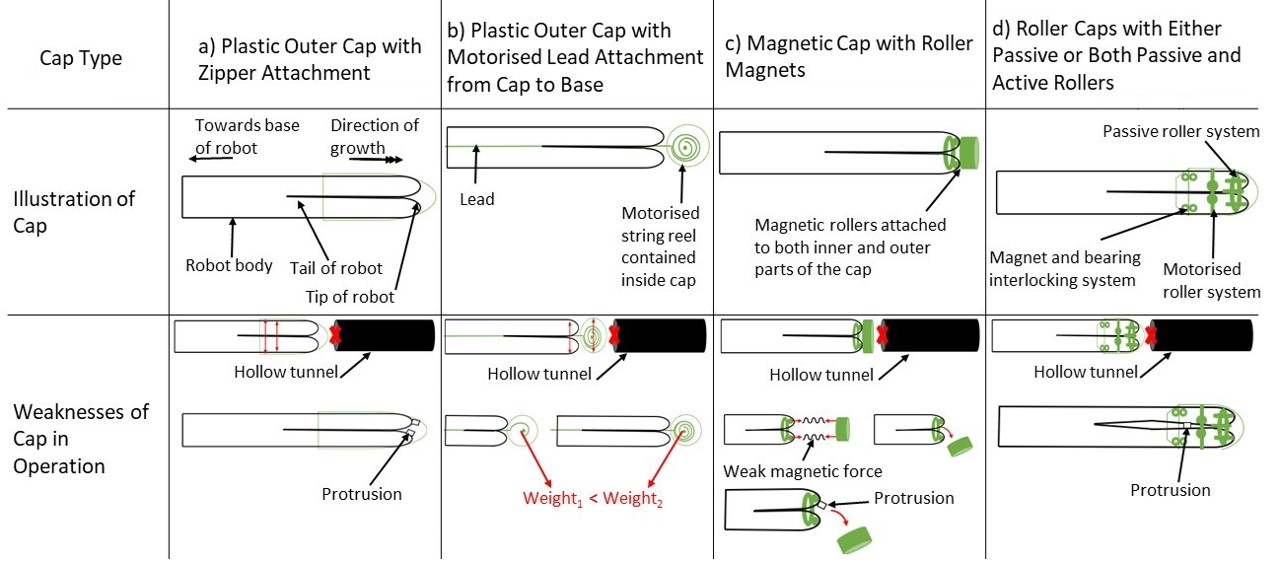}
  \caption{Eversion robot cap designs in literature and their respective weaknesses in certain conditions. a) Plastic outer cap with zipper attachment\cite{Coad2020Vine} b) Plastic outer cap with motorised lead attachment from cap to base \cite{mishima2006development}  c) Magnetic cap with roller magnets \cite{luong2019eversion} d) Cap with roller mechanisms \cite{jeong2020tip, der2021roboa}.}
  \label{fig2}
\end{figure*}


Attaching a tool or sensor to the tip of a vine robot is challenging, and has baffled scientists working in the area - how can one attach a payload to a robot whose tip is constantly evolving \cite{jeong2020tip}?

With this in mind, various cap designs have been suggested, though all suffer from severe shortcomings when compared to our design (Fig. \ref{fig2}). Firstly, their constituent materials are rigid, limiting their ability to squeeze through narrow openings and their capacity to bend their structure. Secondly, interlocking mechanisms linking the cap to the robot tip - whether achieved by rollers or magnets - are mechanically complex, prone to failure and incapable of handling skins with embedded pouches or protruding elements.

In this paper, we present a soft cap (Fig. \ref{fig1}) made from textile material that is able to carry a payload at the tip of an eversion robot. Our soft cap sits on the robot's tip like a beanie and remains there during eversion. Our research shows that a soft cap that is well integrated with the tip of the eversion robot, satisfying essential parameters such as the cap's diameter, conformability and material friction properties, will remain held in position by the sliding motion of the outer skin everting from the tip towards the robot's base. For example, by attaching a wireless camera to the proposed cap atop of an eversion robot, it is possible to inspect environments that would otherwise be unreachable. It is noted that due to the nature of the cap, it does not compromise any of the desired qualities of an eversion robot, such as squeezability and navigability. 


\section{Current Cap Designs}

There are three main functionalities that are important in designing a cap mechanism for an eversion robot: they need to enable rather than hinder movement;  they need to be able to remain fixed at the tip and they need to be able to carry a payload such as a sensor or tool. Some cap designs have  mechanisms to control the length and rate of extension of the robot, while others simply move as the robot inflates. Cap mechanisms seen in the literature that hold the cap at the tip include rollers \cite{jeong2020tip, der2021roboa}, zippers \cite{Coad2020Vine}, leads \cite{mishima2006development}, and magnets \cite{luong2019eversion}. Despite being a drawback for reasons that will become apparent in this section, all eversion robot caps presented in literature at the time of writing this paper are constructed from rigid materials. In this section we will discuss the different designs and outline their operational drawbacks. 


Fig. \ref{fig2} provides diagrammatic representations of four common types of caps and outlines their limitations in certain operating scenarios.

The plastic outer cap \cite{Coad2020Vine}, shown in Fig. \ref{fig2} a), is attached to the robot's body by exploiting the friction between the body and the cap. This friction is usually unwanted as it can inhibit eversion motion, but in this case, is crucial, indeed the key to holding the cap in place. However, the rigid nature of the cap does hinder maneuverability of the tip, and the cap also has limited tolerance of increased friction given its non-elastic properties. This can happen due to changes in robot diameter - itself a consequence of influences from the immediate environment such as the impact of protruding objects - ultimately reducing the application areas of the cap.

The outer cap with motorised lead attachment from cap to base \cite{mishima2006development}, shown in Fig. \ref{fig2} b), is kept at the tip of the robot through precise control of the length of the lead, which supplies power to the cap. This regulation of lead length requires a complex control mechanism.  As in \ref{fig2} a), it also uses a plastic cap, bringing with it the same inherent problems. When, for example, the robot is long, the cap mechanism can run out of storage space for the lead, and/or become too heavy, reducing  the robot's ability to extend and maneuver.


The magnetic cap with roller magnets \cite{luong2019eversion}, shown in Fig. \ref{fig2} c), consists of inner and outer parts that are magnetically attracted to each other by way of magnetic rollers that roll over the robot's body material and enable extension. Any imperfections on the robot's body, such as dirt from the environment or manufacturing imperfections, could cause the magnets to become misaligned or weaken the  attraction between the inner and outer caps sections, potentially causing the cap to fall off. Any areas of thickness in the robot body, caused by additional layers (e.g., integrated navigation pouches) or attached objects (e.g., sensors), would also compromise magnetic attraction. Conversely, if the attraction force between the two magnets is too strong,  the robot's body material may become trapped on account of the frictional force, rendering the robot immobile. 

    
The caps with roller mechanisms \cite{jeong2020tip} (passive and active rollers), and \cite{der2021roboa} (passive rollers), shown in Fig. \ref{fig2} d), are  designs in which the robot body material is fed through rollers, and in which multiple internal and external parts need to work together. The rollers themselves can be passive, reacting purely to the motion of the robot, or actively controlled by a motor. The latter version is the only cap with moving parts that contribute to the process of moving the robot body material. One disadvantage of this system is that if any of the rollers become jammed or the motorised rollers stop working, the entire system stops working. Small changes in robot diameter or the presence of protruding objects are among the potential causes of this kind of breakdown. Another is that the attachment of a sensor or other additional element that changes body thickness \cite{exarchos2022task} is not possible. 
There is another issue here in relation to the motorised rollers. Although they assist in precise control, they can also create vibrations in the system, rendering this option unsuitable for use in fragile or delicate environments. 


Another option that has been suggested is to hold a payload using a tendon \cite{hawkes2017soft}. However, this approach also presents challenges. The everting material can introduce additional friction on the tendon, resulting in a velocity difference between the tip of the eversion robot and the payload. There are some solutions to this problem \cite{new1,new2}, but ultimately, using this method instead of a cap raises concerns about the stability of the payload, particularly when dealing with heavier payloads.


Overall, the caps presented in the literature are well-designed for specific operational situations and tasks and are usually used in controlled settings. The fact, however, that they are made from non-compliant, hard or more rigid materials, means that they compromise the performance of eversion robots, which need to be entirely soft to achieve compliance throughout the robot's structure. A hard cap effectively prevents the robot from being able to move through paths narrower than the cap itself. A cap design based on soft materials, and the consequential compliance it would offer eversion robots, would therefore represent a significant breakthrough, enabling these devices to be properly utilised in a range of environments, a wider variety of operational situations and for a larger number of specific tasks.  

\section{Soft Cap Design}

Here, we take on the challenge to develop such soft cap and design around the shortcomings of its rigid counterparts, taking advantage of the given structural properties of an increasingly popular material for soft robotics: textiles.

\subsection{Friction Design Objectives}

\begin{figure}[h]
  \centering
  \includegraphics[width=1\linewidth]{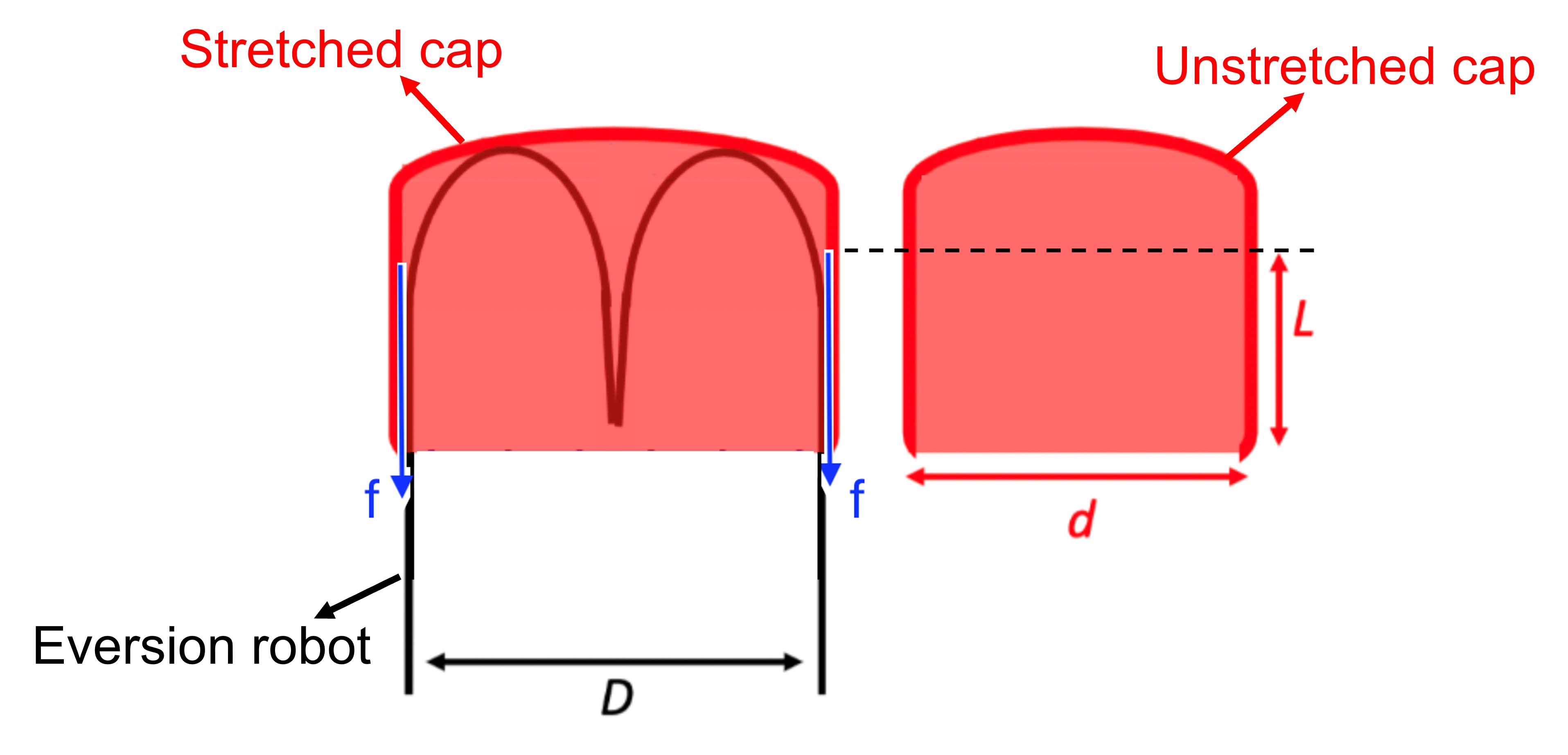}
  \caption{Illustration of exploiting friction between soft cap and downward moving eversion material; (left) an everting eversion robot with a cap mounted; (right) a soft cap in its non-mounted, non-stretched state. The friction force \mbox{\emph{f}} between the cap and the eversion robot body holds the cap in place at the tip. \mbox{\emph{L}} is the length of the effective friction section. \mbox{\emph{D}} is the diameter of the eversion robot as well as the mounted, stretched cap. \mbox{\emph{d}} is the diameter of the unmounted, unstretched cap.}  
  \label{figMI}
\end{figure}

One of the biggest challenges in eversion robot cap design is securing the cap at the tip of the robot. Whether working with hard or soft caps, mounting something onto a moving growing robot creates friction between the everting layers.
The design we introduce here exploits  frictional force between the cap and the eversion robot body as a route to keeping the cap in place without restricting its movement. With the help of  elastic textiles \cite{suulker,sanchez2021textile}, the cap is slipped onto the tip of the everting robot, able to slightly adapt in its diameter through its stretch character, while still encapsulating the body in a firm, yet flexible way - a frictional force enabled by inherent textile properties that we intend to exploit.
More specifically, the typically non-elastic, sturdy and robust material forming the robot's body and everting from the tip of the eversion robot moves, at all times, to the robot's base pulling the sides of the cap with it and ensuring that the cap sits tightly on the tip during eversion (Fig. \ref{figMI}). This frictional force between the elastic fabric and the outer skin of the eversion robot body is essential for the cap to be held in place (Fig. \ref{figMI}). This frictional force is applied in an equally distributed, symmetrical way around the tip, helping to hold the cap's orientation.

As this frictional force poses an advantage for the stable positioning of a cap, it also tends to inhibit the speed of eversion - indeed if the force is too great (the cap too tight around the tip), it prevents motion entirely. A balance must therefore be maintained, that allows eversion while ensuring the cap remains in place. This begs a number of questions such as how narrow the cap can be (with respect to the eversion robot's diameter), or whether a single cap can be used for different diameters of eversion robots? In this paper, we examine these issues. 

There are numerous parameters that can be changed to optimise the effectiveness of the cap - these are illustrated in Fig. \ref{figMI} in which \mbox{\emph{L}} is the length of effective contact surface - the region the friction is created. In this region the cap and the eversion robot are in contact, creating friction that holds the cap in place. \mbox{\emph{D}} and \mbox{\emph{d}} are the diameters of the eversion robot and the (unmounted) cap respectively. $\Delta$\mbox{\emph{D}} is the difference between the two diameters (Equation 1). The key parameter relating to the difference in diameter is \%\mbox{\emph{D}} which can be calculated using Equation 2 and indicates the percentage difference between the two diameters - i.e. \%\mbox{\emph{D}}= 10 means the diameter of the cap is 10\% smaller than that of the eversion robot body.
\begin{equation} \ \Delta{\emph{D}}=\emph{D}-\emph{d} \end{equation}
\begin{equation} \%{\emph{D}}=\dfrac{\Delta \emph{D}}{\emph{D}}\times100 \end{equation}

Existent eversion robots are diverse, though following the same design principle. Parameters that easily change are its diameter, material layers (e.g. through pouches), surface smoothness or unevenness (e.g. through protruding objects), and payloads (e.g. cameras, other sensors), all while retaining the option of passing through the narrowest openings to access target sites. Therefore, a key design objective for engineering a soft cap is to account for these parameters. Under these premises, we have undertaken an iterative, exploratory design journey guided by an in-depth knowledge in textile technology and pattern construction.


\subsection{Prototype Construction}


In this section, the fabrication of a series of prototypes is explained. Changing the diameter \%\mbox{\emph{D}} and length \mbox{\emph{L}} in the designs is also examined.

\begin{figure}[h]
  \centering
  \includegraphics[width=0.9\linewidth]{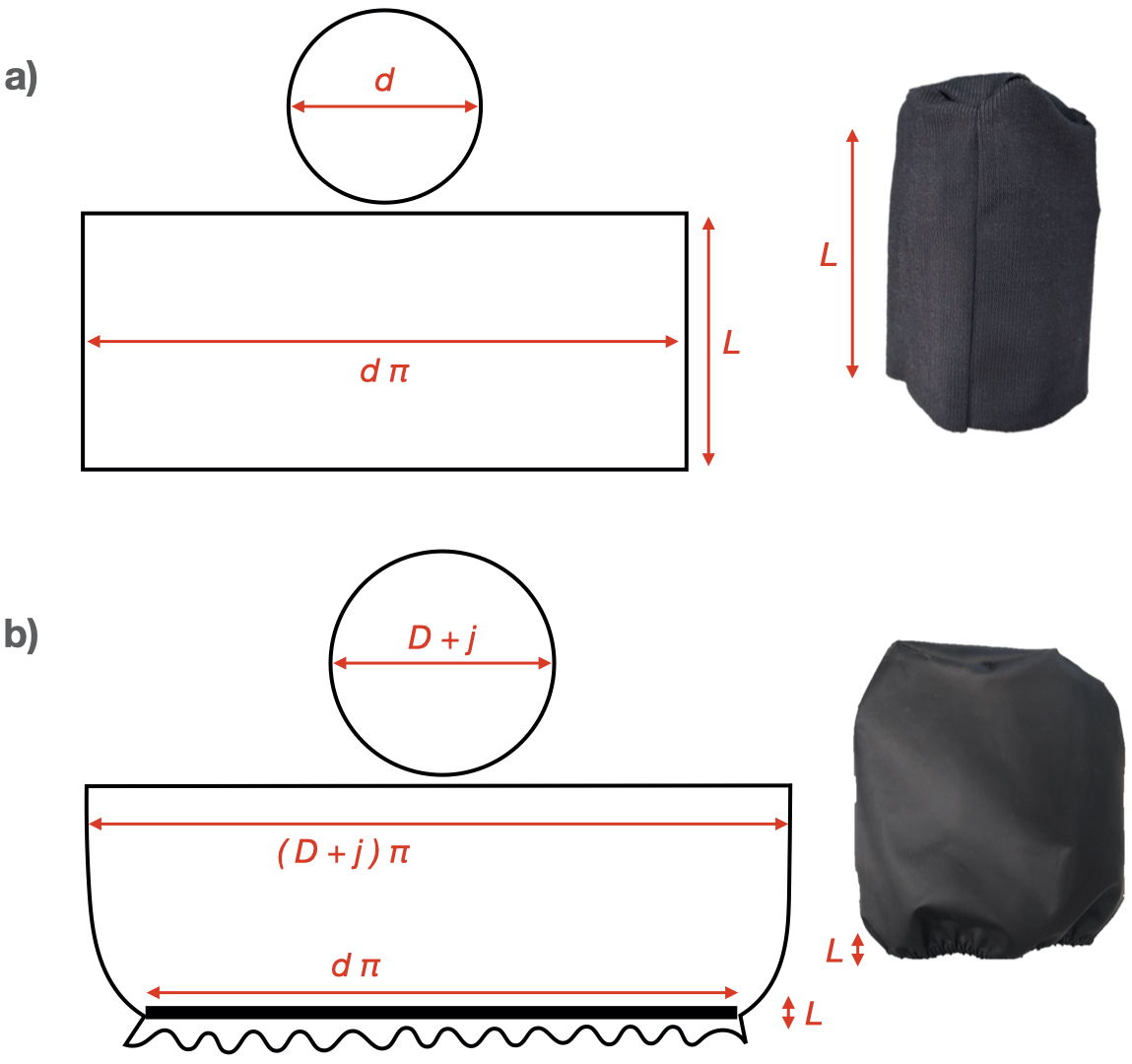}
  \caption{Pattern constructions a) using elastic finely knitted rib fabric material with diameter d, and b) using non-elastic woven fabric and elastic band with diameter D+j, D being the diameter of the robot, and j = 3 cm.}
  \label{figCP}
\end{figure}

With various cutting and sewing patterns that could be explored, this paper focuses on the following approach. A circle is cut from fabric so that it covers the everting tip of the robot; similarly, a rectangle is cut from fabric to cover the outer sides of the robot  (Fig. \ref{figCP}), with the shorter (vertical) edges sewn together to form a cylindrical shape. The two pieces, the circle and the cylinder, are sewn together either by using an `overlocking' machine\footnote{bernette 44 FUNLOCK in 4-thread default settings, producing a more elastic stitch that also cuts and fringe seams the fabric edges.}. In this way, the rounded end of the tube-like cylinder is pinned to the circle and attached section by section, until the rounded shape of a one sided closed cylinder is completed.

To make these textile end caps, two options are considered. First, stretch fabric is used to create the whole cap (Fig. \ref{figCP} a). The diameter of the circular piece is constant for the whole body of the cap and is equal to \mbox{\emph{d}}. The dimensions of the rectangle are \mbox{$\pi\emph{d}\times\emph{L}$}. To ensure secure attachment via a seam, a 1cm seam allowance is added, and later concealed on the inside of the cap. The \mbox{$\pi\emph{d}$} long side is sewn to tangents of the \mbox{\emph{d}} diameter circle. The other sides of the rectangle meet after this step, and they are sewn together to create the cylindrical side wall of the cap.

The elastic fabric is made of cotton mix with elastane yarn integrated to the weft of the fabric (fabric in Fig. \ref{figCP} a). It is woven in the canvas structure, which in absentia is non-stretchy, robust, yet maintains a soft touch. With the embedded elastane, however, additional stretch in weft direction is created, enabling mono-directional expansion. This one way stretch character of the fabric exploited to expand the cap in diameter.

The second method of fabrication requires greater expertise on account of the complexity of working with elastic bands. The cap is made from non-stretch fabric but by using elastic bands, the cap's bottom section will squeeze onto the eversion robot body. An elastic band, also called \textit{elastics}, is commonly used in clothing to create ruffles or elastic waistbands. Being a standard method in the textile industry, it has rarely employed for the development of textile robotics \cite{suulkerral}. It composed of braided polyester and a little part of thin rubber, making it enduring and extremely stretchable. Using this method leads to a smaller effective contact surface \mbox{\emph{L}} (the width of a single elastic band). The circular pattern should be selected bigger than both the effective contact section diameter \mbox{\emph{d}} and robot diameter \mbox{\emph{D}}. The length of the long side of the rectangular pattern, (\emph{D}+\emph{j})\mbox{$\pi$} with \emph{j} being a small number, is matching the diameter of the circular pattern, \emph{D}+\emph{j}. The other side is narrowed by use of a  long elastic band of length, \mbox{$\pi$\emph{d}}. The thickness of the elastic band gives the effective contact length \mbox{\emph{L}}  (Fig. \ref{figCP} b). 
After the same sewing actions are done, the final cap pattern emerges, as can be seen in Fig \ref{figCP} b. The specifications of the prototypes used in this work are summarised in Table \ref{tableprot}.

\begin{table}[h]
\caption{Specifications of the cap prototypes}
\label{tableprot}
\begin{center}
\begin{tabular}{|c||c||c||c|}
\hline
Prototype No  &  Elasticity method & \mbox{\emph{L}} & \%\mbox{\emph{D}} \\
\hline
1 & Elastic fabric & 15 cm & 20\% \\
\hline
\textbf{2} & \textbf{Elastic fabric} & \textbf{15 cm} & \textbf{10\%} \\
\hline
3 & Elastic fabric & 15 cm & 2\% \\
\hline
\textbf{4} & \textbf{Elastic fabric} & \textbf{10 cm} & \textbf{10\%} \\
\hline
5 & Elastic fabric & 5 cm & 10\% \\
\hline
6 & Elastic band & 0.5 cm & 20\% \\
\hline
\textbf{7} & \textbf{Elastic band} & \textbf{0.5 cm} & \textbf{10\%} \\
\hline
\textbf{8} & \textbf{Elastic band} & \textbf{0.5 cm} & \textbf{2\%} \\
\hline
\end{tabular}
\end{center}
\end{table}

\section{Case Study \& Evaluation}


The prototypes whose specifications are listed in Table \ref{tableprot} were exposed to a number of key challenges to evaluate their effectiveness in different environments. These challenges were set up to mimic realistic scenarios, with each one tested on a different eversion robot type. All the tests run 10 times, and percentage-based success criteria are given. The failure for the cap prototype means the eversion robot jams and stops everting at some point for challenges 1-4. 

Also for comparison, two rigid caps were built and exposed to the challenges. The first one is labeled as "Rigid Tube Cap" similar to \cite{Coad2020Vine}. It is a 30 cm long cylindrical tube that has one closed end. The diameter is 0.5 cm larger than the eversion robot. The second one is labeled "Rigid Roller Cap", similar to \cite{jeong2020tip}, with passive rollers.

\subsection{Challenge 1: Eversion of many layered bodies}

The maneuvering mechanisms of eversion robots often require additional layers of materials on the  robot body, i.e., navigation pouches \cite{abrar2021highly}, layer jamming elements \cite{exarchos2022task}. These extra layers change the wall thickness and the diameter of the robot, creating significant challenges for caps, especially those made from rigid material. Changes in wall thickness can cause jamming in rigid caps that use rollers and  contact loss in those that use magnets.  This is because such caps are generally custom-built for a specific diameter, and therefore not robust enough to compensate for this kind of change. However, the problem is even greater for friction-based systems as these changes create dramatic variations in the frictional force between the cap and outer eversion robot skin.  

\begin{figure}[h]
  \centering
  \includegraphics[width=0.8\linewidth]{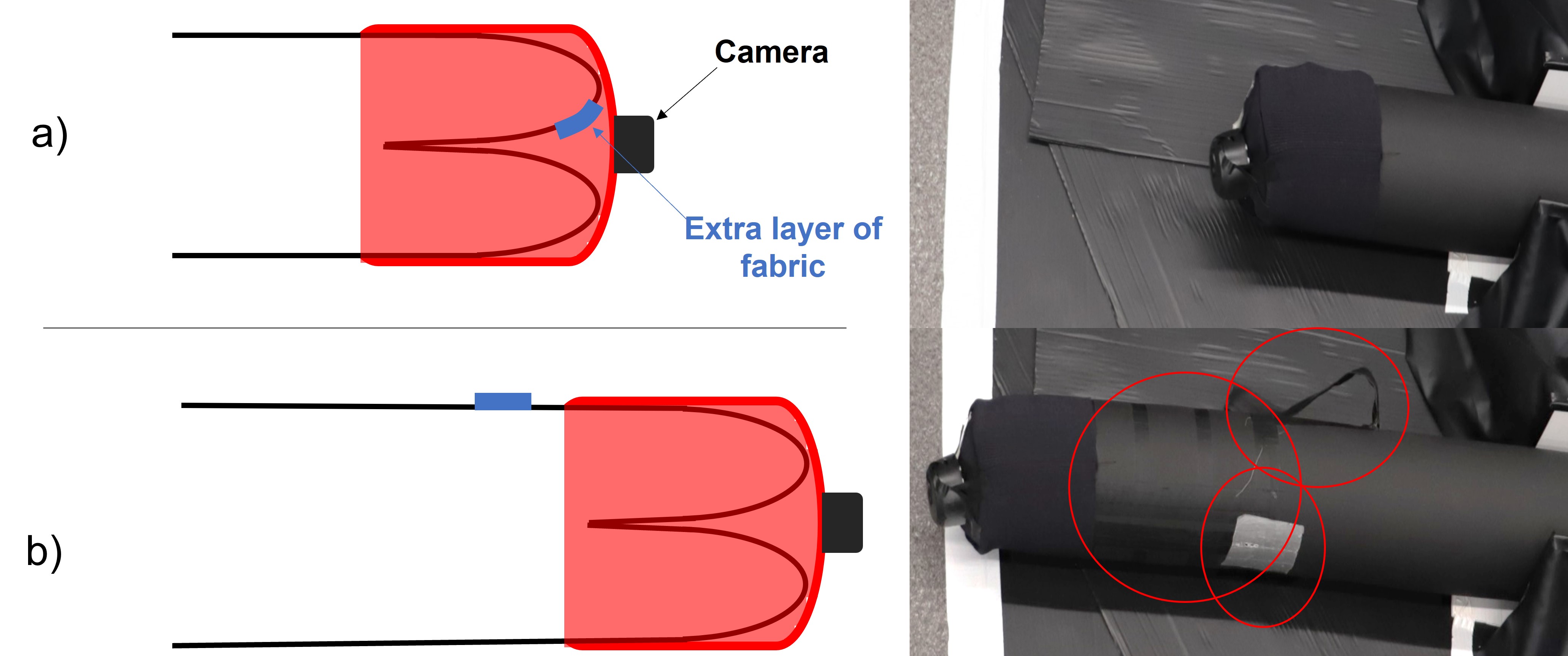}
  \caption{ a) An eversion robot with soft cap (Prot. 4) and a camera attached to it. b) It successfully completing "Challenge 1". Additional layers of materials that change the thickness of the body are marked.}
  \label{figCh1}
\end{figure}

For this challenge we attached four sets of pouches to our eversion robot body, increasing the thickness of the robot's walls by 2\% (0.2 cm) and its diameter by 4\% (0.4 cm). To increase the thickness even more, duct tape and loose fabric were attached to the robot body. Overcoming this challenge would prove that the cap could be used in eversion robots that have thick walls and change diameter along their length.
In the starting condition (Fig. \ref{figCh1} a) the robot is inflated and extra layers are not everted yet. To successfully complete the challenge the robot with a cap should grow despite the extra layers, and the layers should be fully visible leaving the boundaries of the cap (Fig. \ref{figCh1} b).

\subsection{Challenge 2: Eversion of protruding solid objects}

This challenge simulates the placement of sensors, electrical components, air tubes and rigid connectors on the body of the eversion robot. As with, or perhaps even more so than with extra layers, these rigid components can cause jamming in caps with complex mechanisms,  entirely blocking movement in the robot. 

\begin{figure}[h]
  \centering
  \includegraphics[width=0.8\linewidth]{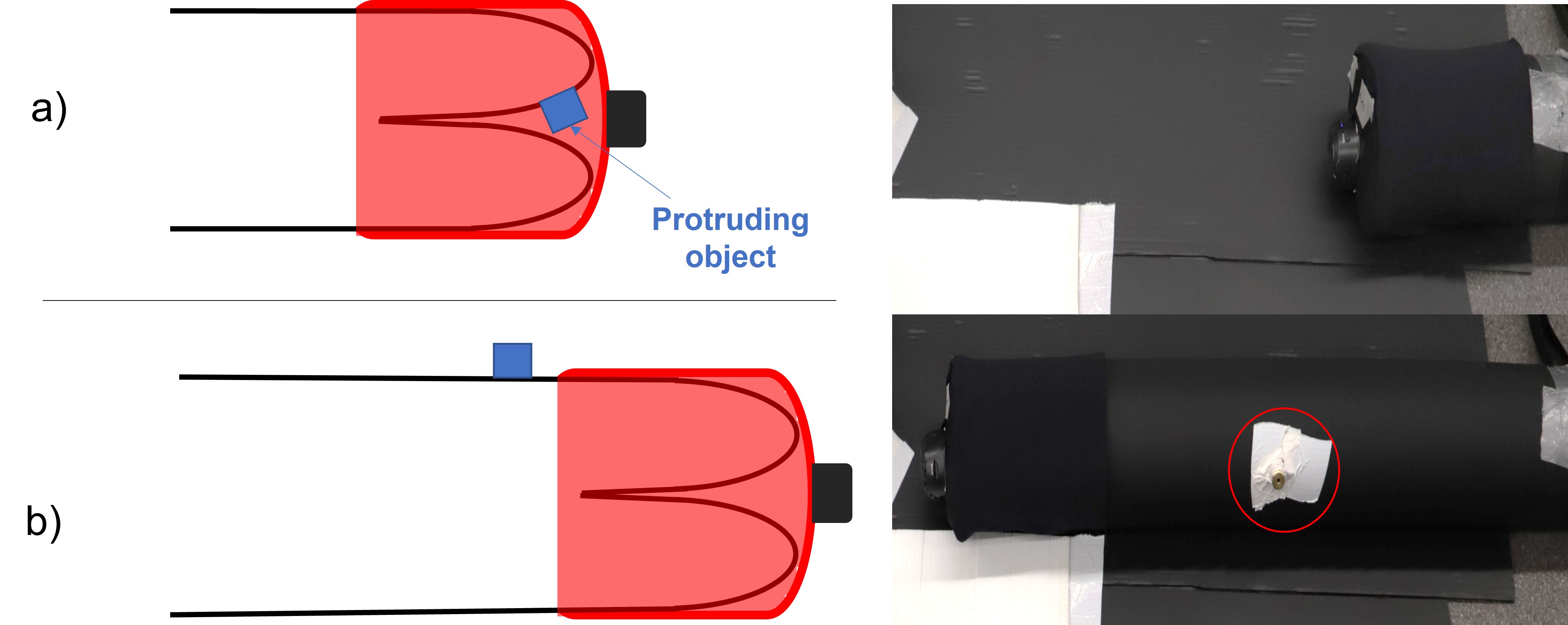}
  \caption{a) An eversion robot with a soft cap (Prot. 2) and a camera attached to it b) It successfully completing "Challenge 2". 1.7 cm long protruding material is circled.}
  \label{figCh2}
\end{figure}

To simulate this, we attach two 1 cm wide, 1.3 cm long and 1.6 cm wide, 1.7 cm long pipe connectors to the robot body in random places. They are roughly taped to the robot body to heighten the difficulty of the challenge. 
In the starting condition the robot is inflated and protruding solid objects are not everted yet (Fig. \ref{figCh2} a). For successfully completing the challenge the robot with a cap should grow despite the protruding objects, and the objects should be fully visible leaving the boundaries of the cap (Fig. \ref{figCh2} b).

\subsection{Challenge 3: Squeezability}

The ability of soft eversion robots to squeeze through narrow openings has been marketed as one of their most alluring properties \cite{hawkes2017soft}. However, a rigid cap mounted at the tip significantly compromises  this function. With our soft cap approach, we  ensure this property is retained, whilst also extending the robot's capability to carry a payload at its tip.

\begin{figure}[h]
  \centering
  \includegraphics[width=1\linewidth]{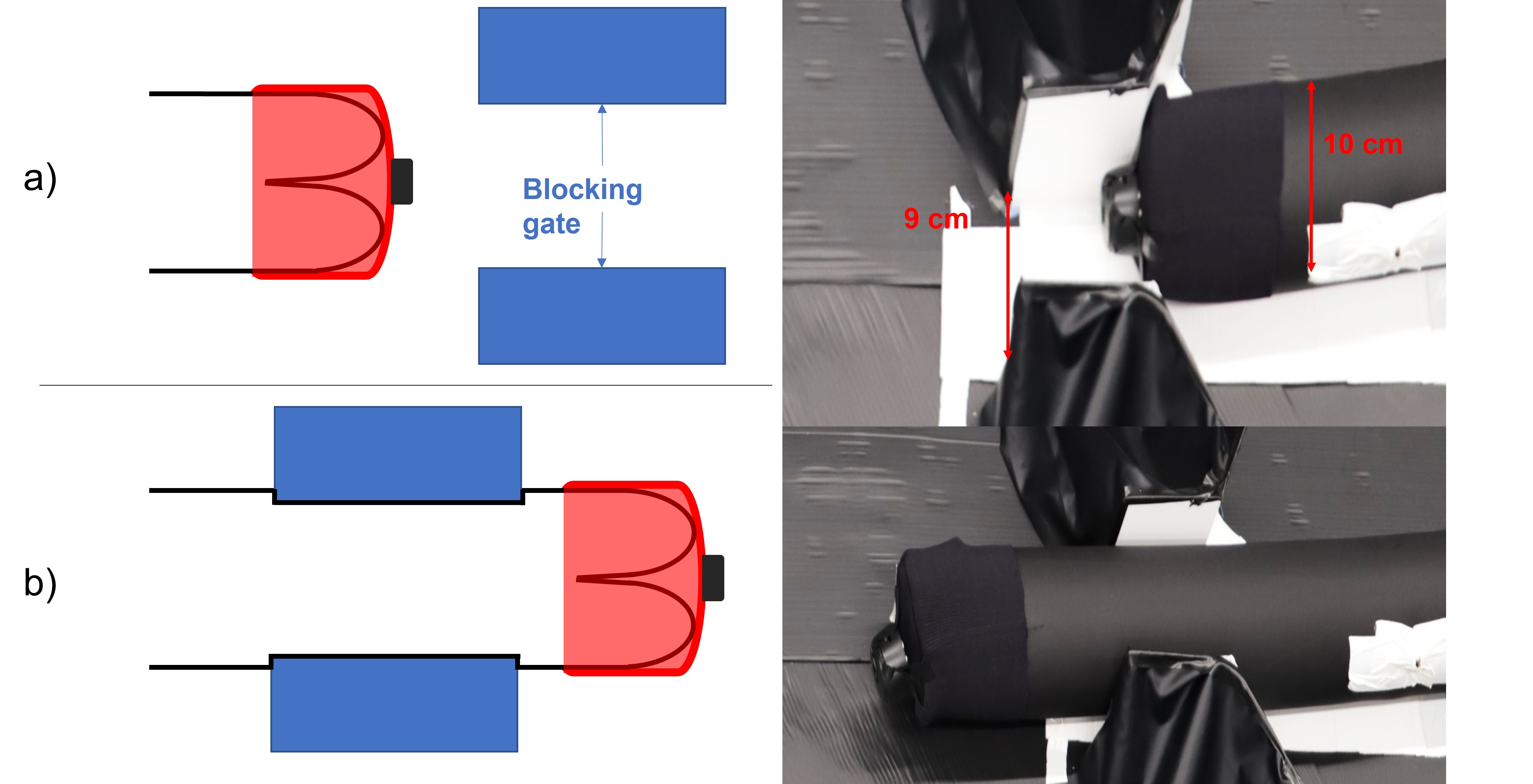}
  \caption{An eversion robot with soft cap (Prot. 4) and camera attached to it, successfully completing "Challenge 3". Diameter of the robot and width of the opening are indicated.}
  \label{figCh3}
\end{figure}

In this challenge, we built a 9 cm wide gate for a 10 cm diameter eversion robot. The robot with the soft cap is able to pass through the narrow opening and continue its path. To prove our concept, we opted for a gate  10\% smaller than the body of the robot.
The starting condition of the challenge is; the eversion robot with a payload attached cap gets inflated 15 cm away from the gate (Fig. \ref{figCh3} a). If the robot can pass through and 20 cm beyond the gate, with the cap remaining on the tip, the challenge is considered successful (Fig. \ref{figCh3} b).


\subsection{Challenge 4: Navigability}

Rigid caps restrict the expansion of the eversion robot's navigation mechanisms severely, leading to reduced tip mobility. A soft cap can tolerate greater expansion without compromising mobility.

\begin{figure}[h]
  \centering
  \includegraphics[width=0.9\linewidth]{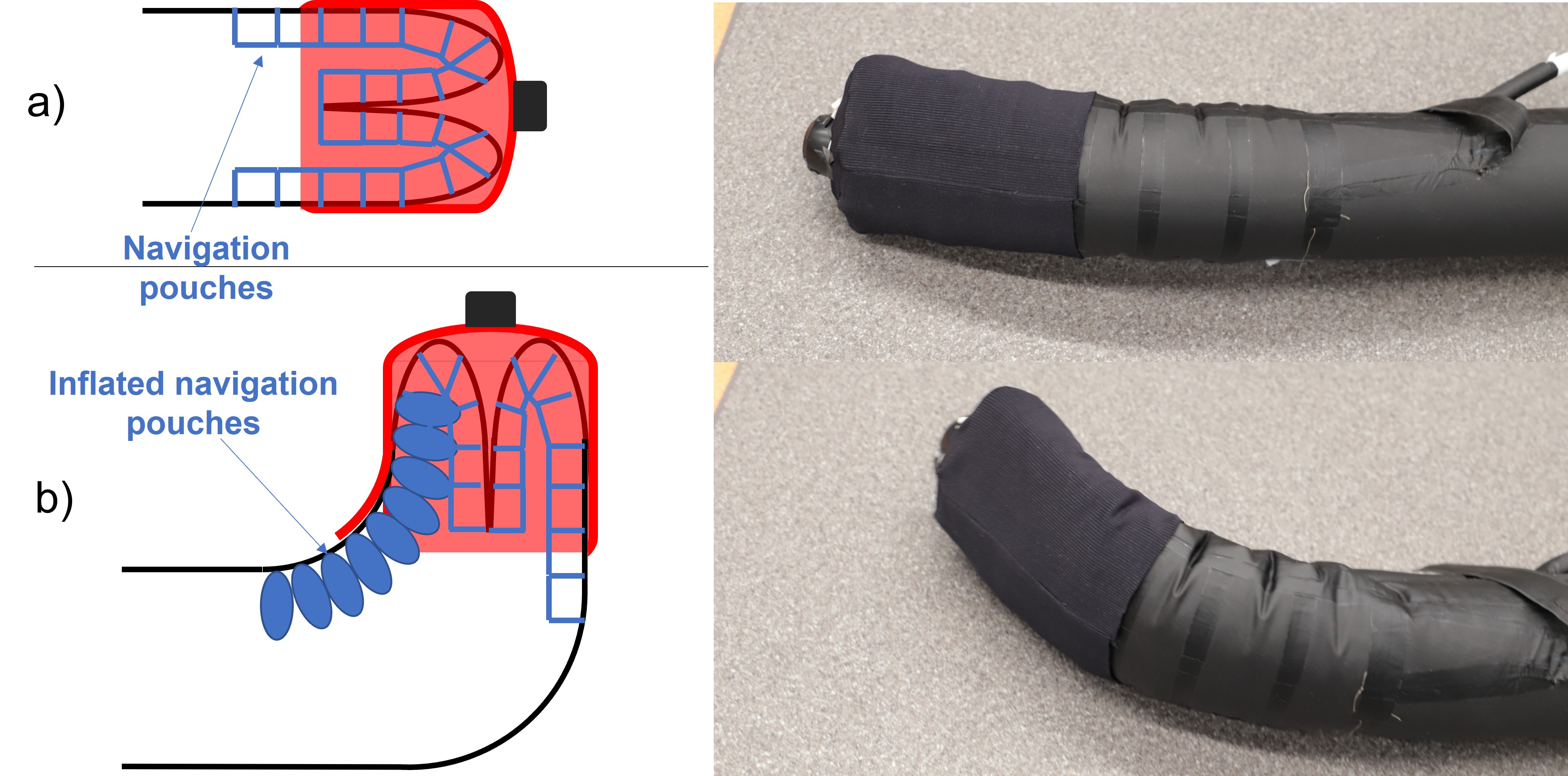}
  \caption{An eversion robot with a soft cap (Prot. 3) reorienting its tip. It is halfway completing "Challenge 4".}
  \label{figCh4}
\end{figure}

In this challenge, we therefore activate our eversion robot with cap in situ, the challenge being to retain the cap at the tip while allowing requisite motion. At starting position the cap (Fig. \ref{figCh4} a) is attached to the eversion robot which is the halfway through eversion of the navigation layers (pouches \cite{abrar2021highly}), and then first the navigation layers then the main chamber of the robot are being inflated. If the tip of the robot embraces the orientation (Fig. \ref{figCh4} b) change due to the navigation mechanism and keeps growing until it passes the navigation layers, the prototype is considered successful. 

\subsection{Challenge 5: Payload Stability}

The payload stability is not a challenge for most rigid caps. However, for the soft cap, we observed that often the attached payload points different directions after the challenges are realised, especially challenge 3, squeezability. For vision feedback payloads like a camera, this is critical merit.

\begin{figure}[h]
  \centering
  \includegraphics[width=1\linewidth]{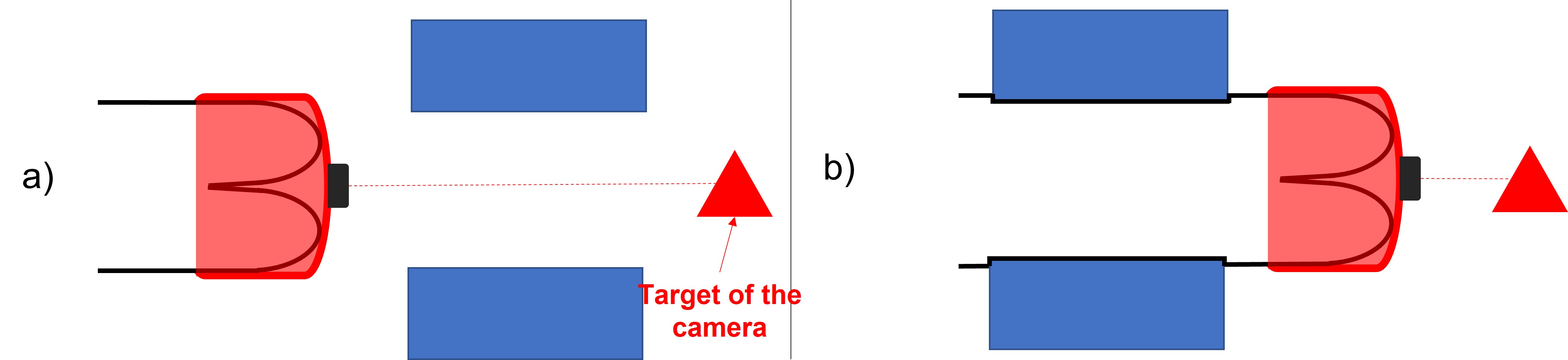}
  \caption{a) Starting condition of the "Challenge 4" b) Success condition for the challenge.}
  \label{figCh5}
\end{figure}

The starting condition of the challenge is the eversion robots start being inflated 15 cm away from the narrow gate from challenge 3, the camera pointing to the centre of an equilateral triangle object (5 cm edge length) which is 2 m away from the gate (Fig. \ref{figCh5} a). After the robot passes through and 20 cm beyond the narrow gate if the camera still sees the centre of the triangular object (Fig. \ref{figCh5} b), it succeeds in the challenge, otherwise it fails. Hence the relevant cap prototype can only succeed in this challenge if it did in challenge 3.

\section{Results \& Discussion}

The fabricated prototypes listed in Table \ref{tableprot} were subjected to each of the aforementioned challenges. Additionally, a wireless camera was deployed at the centre tip of the caps to assess the payload stability during each test. An overview with the results of all experiments are listed in Table \ref{tablechal}.


\begin{table}[h]
\caption{Success percentage of each prototype confronted with key challenges while carrying a payload}
\label{tablechal}
\begin{center}
\resizebox{0.48\textwidth}{!}{\begin{tabular}{|c||c||c||c||c||c||c|}
\hline
Prot. &  Chal. 1 & Chal. 2 & Chal. 3 & Chal. 4 & Chal. 5 & Overall \\
\hline
1 & 100\% & 100\% & 30\% & 0\% & 80\% & 62\%\\
\hline
\textbf{2} & \textbf{100\%} & \textbf{100\%} & \textbf{100\%} & \textbf{80\%} & \textbf{50\%} & \textbf{86\%}\\
\hline
3 & 100\% & 100\% & 100\% & 100\% & 20\% & 84\%\\
\hline
\textbf{4} & \textbf{100\%} & \textbf{100\%} & \textbf{100\%} & \textbf{100\%} & \textbf{80\%} & \textbf{96\%}\\
\hline
5 & 0\% & 0\% & 0\% & 0\% & N/A & 0\%\\
\hline
6 & 100\% & 100\% & 20\% & 50\% & 40\% & 62\%\\
\hline
\textbf{7} & \textbf{100\%} & \textbf{100\%} & \textbf{100\%} & \textbf{100\%} & \textbf{90\%} & \textbf{98\%}\\
\hline
\textbf{8} & \textbf{100\%} & \textbf{100\%} & \textbf{100\%} & \textbf{100\%} & \textbf{50\%} & \textbf{90\%}\\
\hline
Rigid & & & & & & \\ Tube Cap & 100\% & 0\% & 0\% & 0\% & N/A & 20\%\\
\hline
Rigid & & & & & & \\ Roller Cap & 0\% & 0\% & 0\% & 0\% & N/A & 0\%\\
\hline
\end{tabular}}
\end{center}
\end{table}

At first glance at Table \ref{tablechal}, we see that all but one prototype perform reasonably well and master most challenges while maintaining a stable position of the camera. Caps 2, 3, 4, 7 and 8 are able to satisfy every challenge. In the supplementary video, prototype 3 can be seen to be satisfying the challenges in a single run. 
The soft cap prototype that has failed all tests (5 on the list) is the shortest with only 5 cm in length and is made from stretch fabric.

Looking at the results challenge by challenge, we can further evaluate the other prototypes. When everting multiple layers of the soft robotic body (challenge 1), we find that a too short cap (prototype 5) is problematic, while there does not seem to be any advantage over one of the fabric choices (elastic knit fabric or non-stretch fabric with elastic band) in general. 
Even more forgiving with the different design parameters seems to be the challenge of everting protruding objects between the cap and the body surfaces. All soft fabric caps we produced except cap 5, are able to process such obstacles and adapt in form and stretch capacity as needed. 
It becomes more tricky with challenge 3, `squeezing' the robot through narrow paths, where beside cap 5, also prototype 1 and 6 have low success rate. These are the prototypes with the longest cap length and the highest stretch factor. It appears that for challenge 4, when actuating joint-like pouches, the larger the surface creating friction, the worse this form of navigability. Cap 1 with 15 cm length of stretch fabric and 20\% \%\mbox{\emph{D}} hinders performance. We can see, however, that the same types of caps just a few cm shorter, prototype 4, or slightly less tight, prototype 2, succeed easily.

What do these findings tell us about the design engineering differences of the 8 presented fabric caps? In summary, we can observe that the caps that sit tightest around the robot are most robust and stable in regards to payload positioning, but can be too tight for other challenges. 
On the other hand, we can see from prototype 1 that if the diameter difference between the stretch fabric cap and the eversion robot is too large, the design fails to carry out the tasks. This is mainly due to high friction between the body and the cap preventing smooth eversion.
Similarly, it is also clear that when a cap is too short and too loose, all potential obstacles push it off the tip of the robot body too easily and no other maneuvering is possible.
In general, caps made from knitted stretch fabric achieve high performance with parameters of about 10\% \%\mbox{\emph{D}} and 10-15 cm \emph{L}. For elastic band designs 2-10\% \%\mbox{\emph{D}} and 0.5 cm \emph{L} (width of a single elastic band) offer better results.

As expected, the rigid caps had very low success rate of completing these challenges. Failing the squeezability challenge for both caps was as expected. They also failed protruding objects challenge because the objects jammed between the eversion robot and the rigid components of the caps, preventing the movement of the robot. In a similar fashion, expanding of the pneumatic navigation pouches jammed the rigid caps. The rigid tube cap succeeded in the additional layers challenge since there was enough tolerance for additional fabric to move through, but for the roller cap, these layers entangled on the rollers and caused the failure. This shows that our soft caps can achieve tasks and reach environments that their solid counterparts cannot.







    

This paper presents the first soft cap for eversion robots. 
Exploiting textile properties, the cap adjusts to the everting robot and sits firmly, yet flexible at the tip of the robot's body throughout the eversion process. 
This novel design preserves squeezability and navigability of the robot, and the eversion of thick robot wall layers as well as those with protruding elements is shown to be achievable.

The results of our experiments provide design guidance for further developments of soft caps. Depending on application areas, if payload is needed and environments and pathways are challenging, then the strategy of using a tight, but long cap appears most successful. 

While our design masters key challenges eversion robots are confronted with, a limitation that remains, the retrieval of the cap, is the subject of future work.
Further, the design parameters and challenges tested here cover key aspects of soft eversion robots, but there is more to explore. Experimenting with various cap pattern constructions, variations of stretch, and testing the limits of cap sizes and payloads are future tasks we develop based on the success of this first design of a soft textile cap for eversion robots.

\section*{ACKNOWLEDGMENT}

The authors thank Mish Toszeghi, Rodrigo Zenha, Abu Bakar Dawood and Hassan Mirza for their valuable help. Thanks also to the reviewers for their comments.

\bibliographystyle{IEEEtran} 

\bibliography{references}
\vspace{12pt}

\end{document}